\documentclass[runningheads]{llncs}
\usepackage{graphicx}
\usepackage{amsmath,amssymb} 
\usepackage{color}\usepackage[width=122mm,left=12mm,paperwidth=146mm,height=193mm,top=12mm,paperheight=217mm]{geometry}

\usepackage{epstopdf}
\usepackage{epsfig}
\usepackage{amsmath}
\usepackage{amssymb}
\usepackage{algorithmic}
\usepackage{algorithm}
\usepackage{multirow}
\usepackage{textcomp}
\usepackage{xspace}
\usepackage{color}
\usepackage[tight,scriptsize]{subfigure}
\usepackage{wrapfig}
\usepackage{lipsum}
\usepackage{array}
\usepackage{enumitem}

\def \I {\mathbf{I}}

\def \0 {\mathbf{0}}
\def \1 {\mathbf{1}}

\makeatletter
\DeclareRobustCommand\onedot{\futurelet\@let@token\@onedot}
\def\@onedot{\ifx\@let@token.\else.\null\fi\xspace}

\def\eg{\emph{e.g}\onedot} 
\def\ie{\emph{i.e}\onedot} 
 
\def\etc{\emph{etc}\onedot} 
 
\def\etal{\emph{et al}\onedot} 

\newcommand{\header}[1]{\smallskip\noindent\textbf{#1}}

\newcolumntype{R}[1]{>{\raggedright\arraybackslash\hspace{0pt}}m{#1}}

\begin{document}
\pagestyle{headings}
\mainmatter
\title{Human Centred Object Co-Segmentation} 

\titlerunning{Human Centred Object Co-Segmentation}

\authorrunning{Chenxia Wu, Jiemi Zhang, Ashutosh Saxena, Silvio Savarese}

\author{Chenxia Wu\inst{1} \and Jiemi Zhang\inst{2} \and
Ashutosh Saxena\inst{1} \and Silvio Savarese\inst{3}}

\institute{Computer Science Department, Cornell University,\\
\email{chenxiawu@cs.cornell.edu}, \email{asaxena@cs.cornell.edu} \and
Didi Research,
\email{jmzhang10@gmail.com} \and
Computer Science Department, Stanford University,
\email{ssilvio@stanford.edu}
}

\maketitle


\begin{abstract}
Co-segmentation is the automatic extraction of the common semantic regions given a set of images. Different from previous approaches mainly based on object visuals,
in this paper, we propose a human centred object co-segmentation approach, which uses the human as another strong evidence. In order to discover the rich internal structure of the objects reflecting their human-object interactions and visual similarities, we propose an unsupervised fully connected CRF auto-encoder incorporating the rich object features and a novel human-object interaction representation. We propose an efficient learning and inference algorithm to allow the full connectivity of the CRF with the auto-encoder, that establishes pairwise relations on all pairs of the object proposals in the dataset. Moreover, the auto-encoder learns the parameters from the data itself rather than supervised learning or manually assigned parameters in the conventional CRF. In the extensive experiments on four datasets, we show that our approach is able to extract the common objects more accurately than the state-of-the-art co-segmentation algorithms.



\end{abstract}

\section{Introduction}\label{sec:intro}
Image co-segmentation is defined as automatically extracting the common semantic regions, called \emph{foregrounds}, given a set of images~\cite{Rother_CVPR_2006,Hochbaum_ICCV_2009,Batra_IJCV_2011,Fu_CVPR_2015}. It provides an unsupervised way to mine and organize the main object segment from the image, which is useful in many applications such as object localization for assistive robotics, image data mining, visual summarization, \etc.

\begin{figure}[t]
  \begin{center}
  \subfigure[Original image]{
  \begin{minipage}{0.3\linewidth}
  \includegraphics[width=1\linewidth]{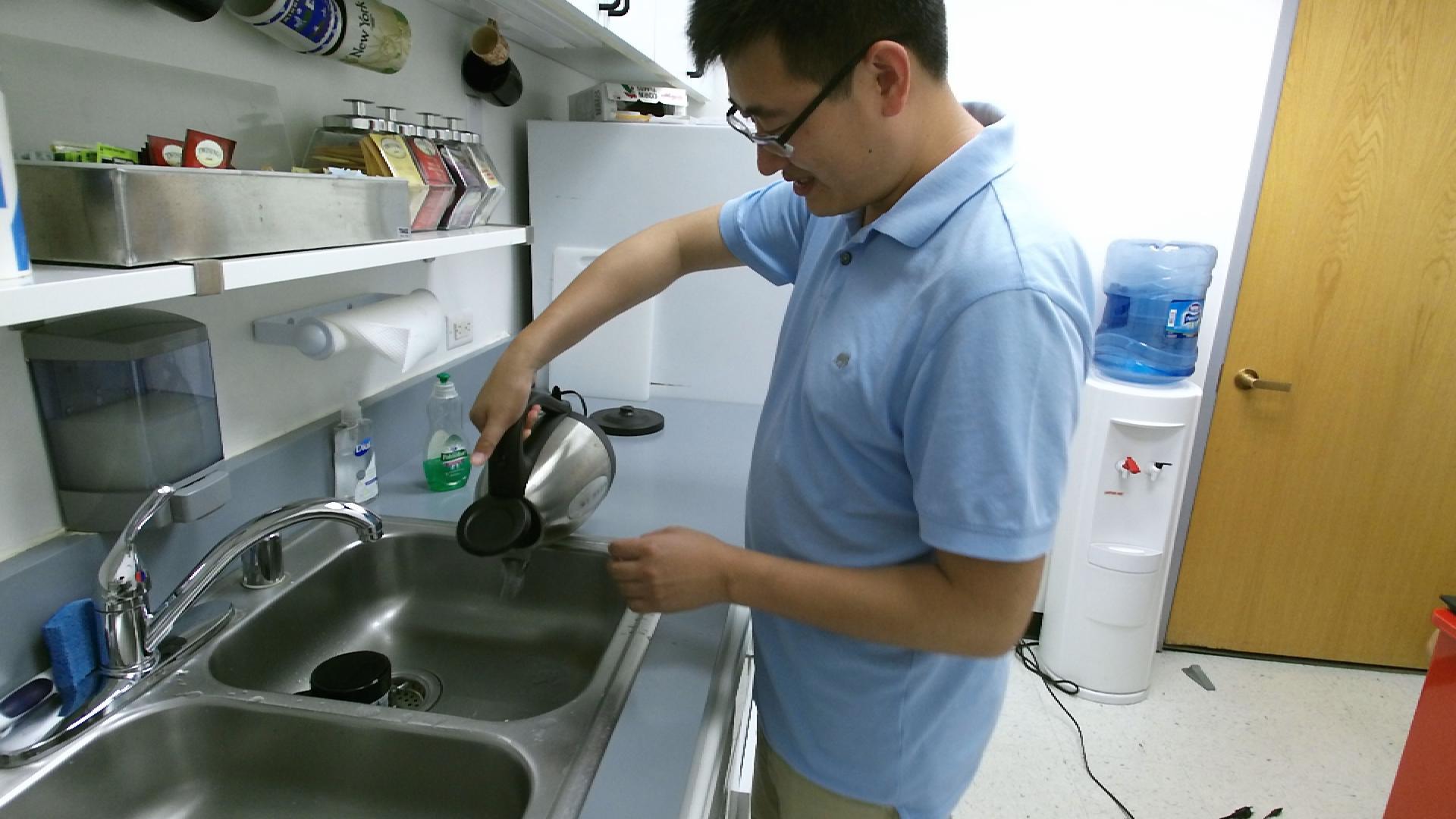}\vspace{0.05in}\\
  \includegraphics[width=1\linewidth]{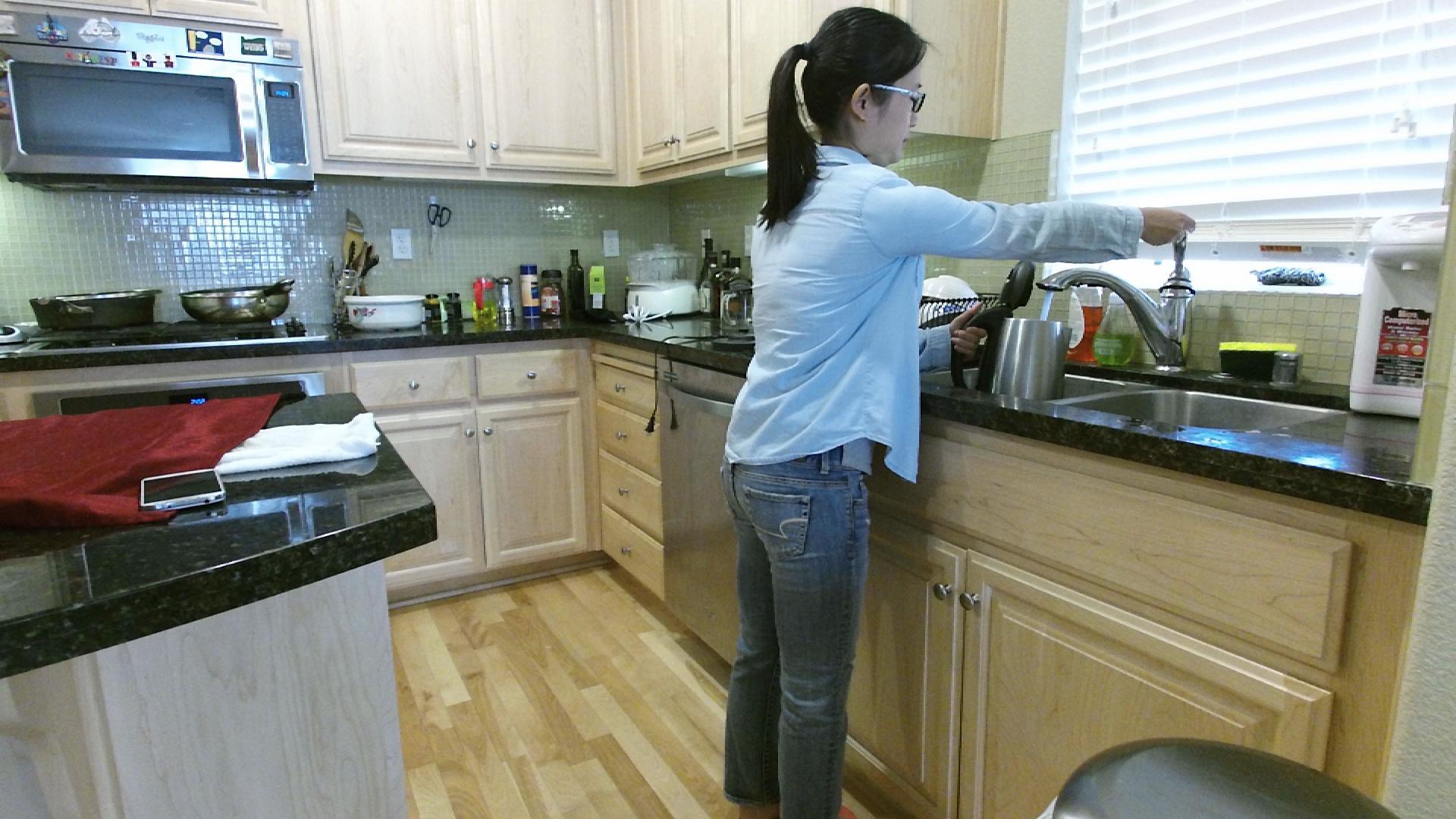}
  \end{minipage}
  }
  \subfigure[Co-segmentation results using only object appearance]{
  \begin{minipage}{0.3\linewidth}
  \includegraphics[width=1\linewidth]{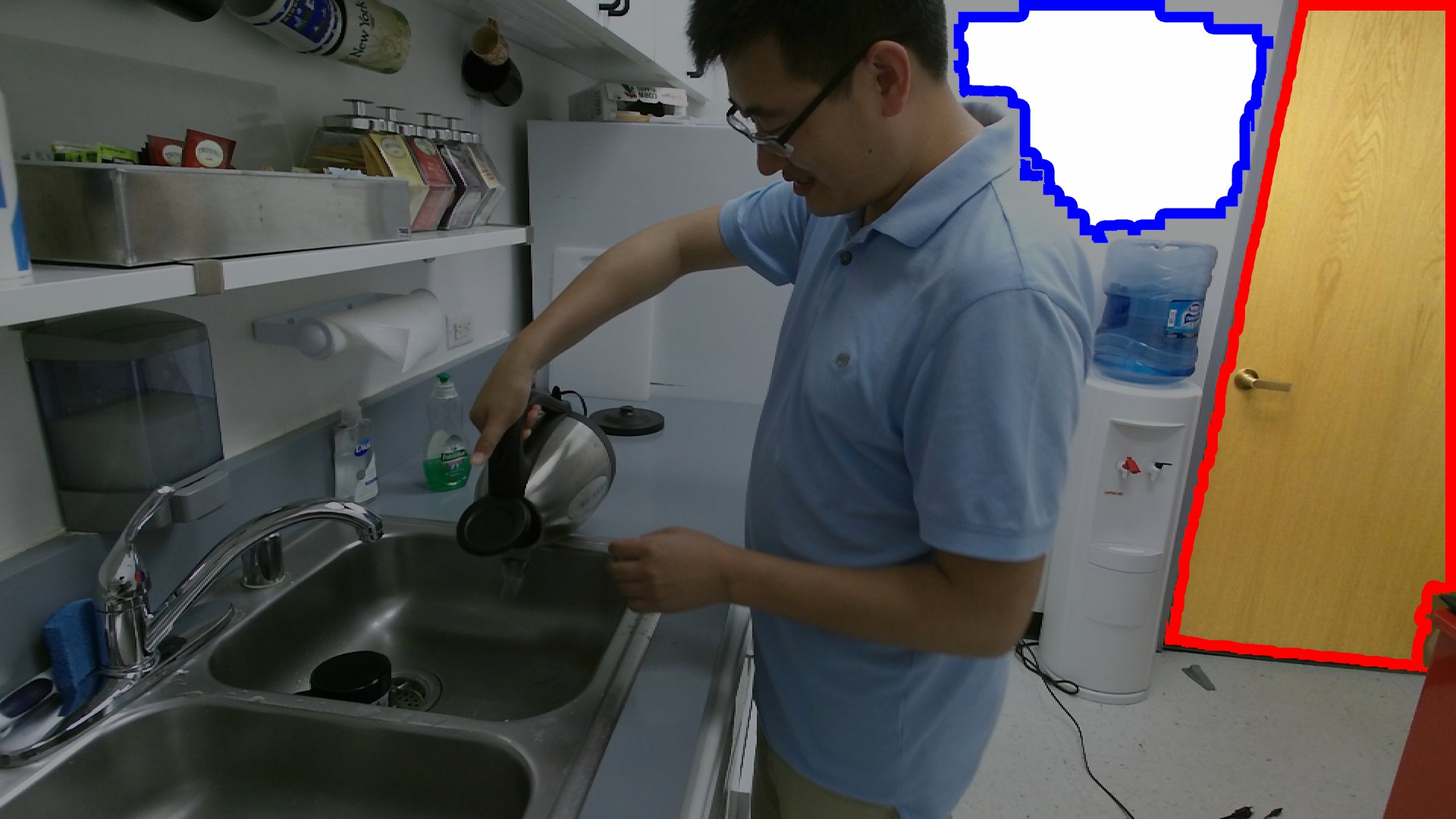}\vspace{0.05in}\\
  \includegraphics[width=1\linewidth]{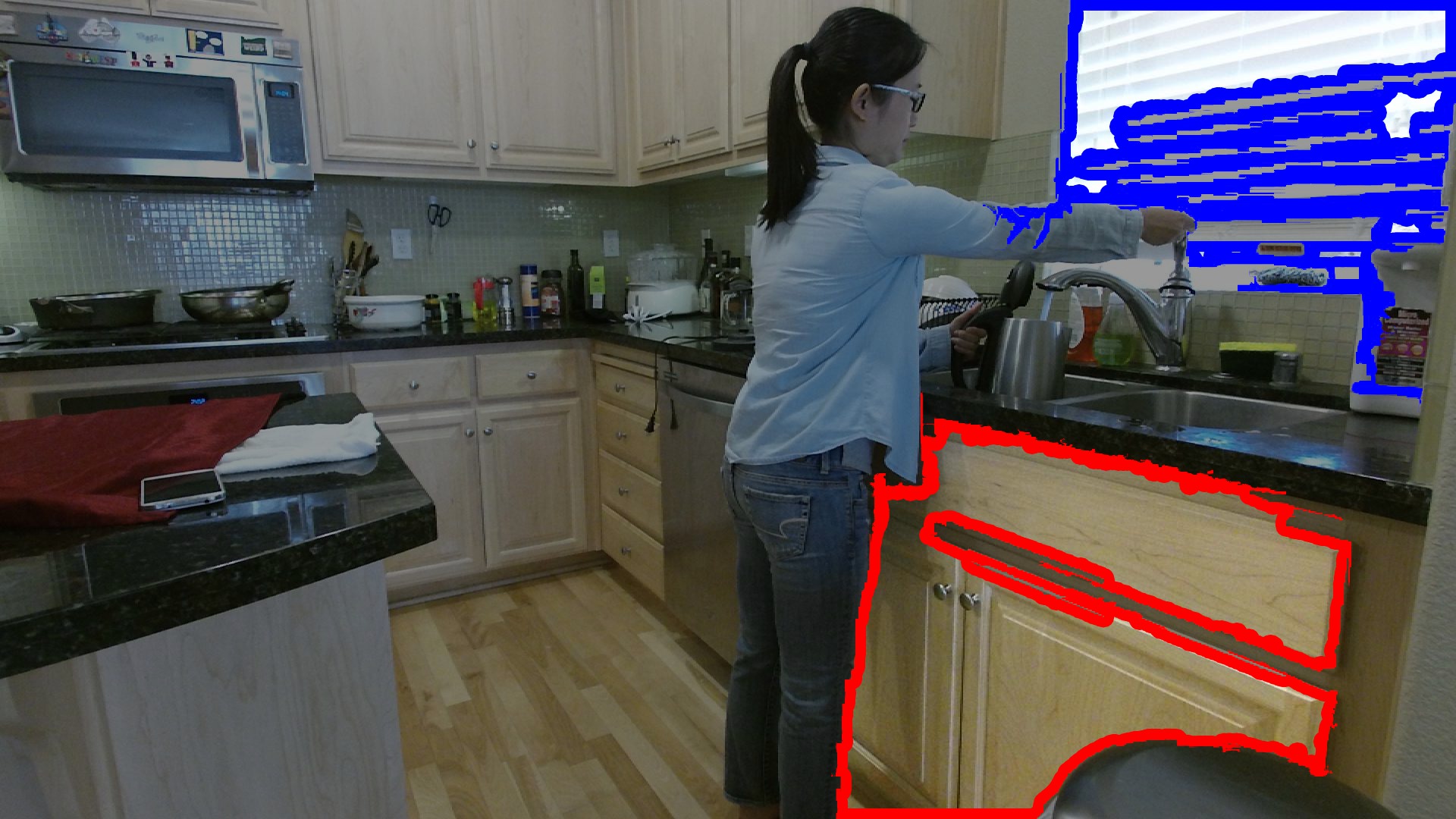}
  \end{minipage}
  }
  \subfigure[Co-segmentation results using object appearance and human]{
  \begin{minipage}{0.3\linewidth}
  \includegraphics[width=1\linewidth]{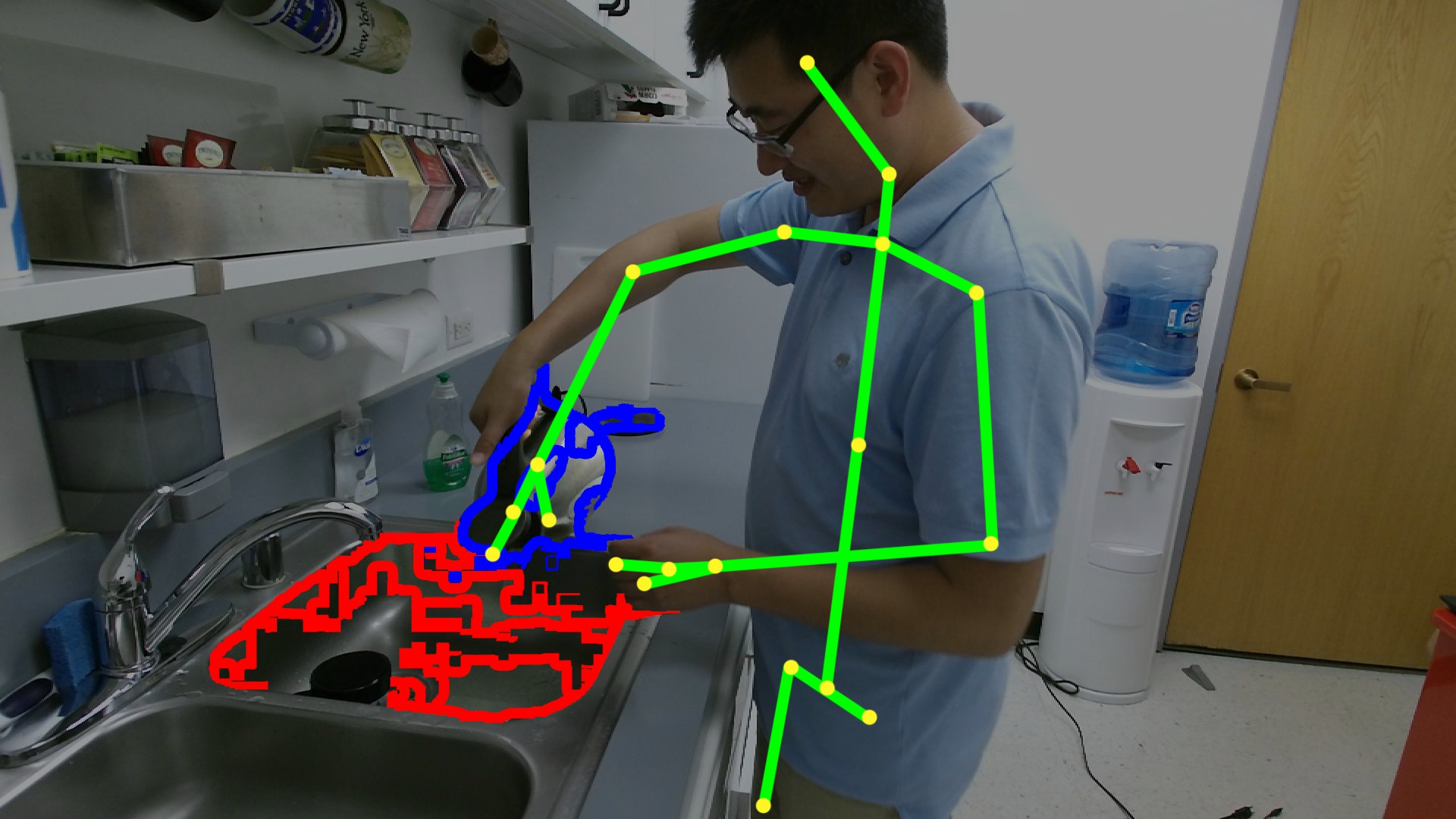}\vspace{0.05in}\\
  \includegraphics[width=1\linewidth]{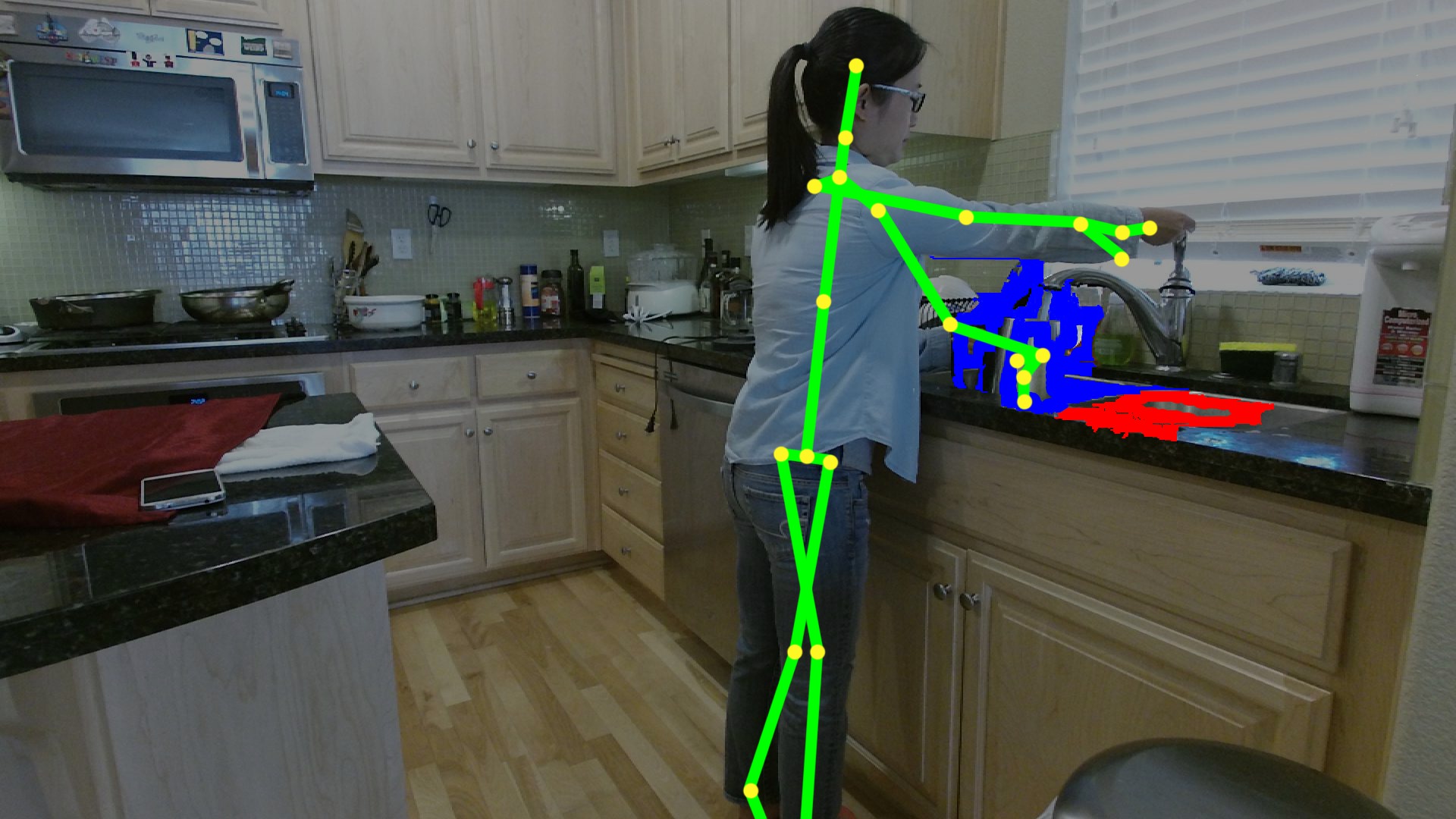}
  \end{minipage}
  }
  \vspace{-0.15in}
 \caption{We propose a human centred object co-segmentation approach by modeling both object visual appearance and human-object interactions.  As in the example, human-object interactions help mining of more useful objects (the pots and the sinks) more accurately in the complex backgrounds, with view changes and occlusions. (Output foregrounds are blue, red circled and human skeletons are green colored.)} \label{fig:task}
 \end{center}
 \vspace{-0.35in}
\end{figure}


The challenge of co-segmentation is that semantic labels are not given and we only have the visual information from the images. The only assumption is that the given images share some common semantic regions, \ie, they belong to the same semantic category. So far, the dominating approaches are to \mbox{co-discover} these common regions only relying on their visual similarities~\cite{Fu_CVPR_2015,Vicente_CVPR_2011,Meng_TMM_2012,Joulin_CVPR_2012}. For instance, \cite{Hochbaum_ICCV_2009,Vicente_ECCV_2010,Batra_CVPR_2010} propose methods for unsupervised pixel-accurate segmentation of ``similarly looking objects'' in a given set of images. Vicente \etal~\cite{Vicente_CVPR_2011} then introduce the concept of ``objectness'', which follows the principle that the regions of interest should be ``objects'' such as bird or car, rather than ``stuff'' such as grass or sky. This helps focus the ``attention'' of the co-segmenter to discover common objects of interest as opposes to irrelevant regions.


However, we argue that in situations where either objects' appearance changes due to intra-class variations are severe (the sinks in Fig.~\ref{fig:task}) or when objects are observed under large view changes or partially occluded by other objects or humans (the pots in Fig.~\ref{fig:task}) or when multiple “salient” objects are present in the image (the door and the counter are also salient in Fig.~\ref{fig:task}), existing co-segmentation methods will not work well. In this paper, we argue that when images do contain humans that use objects in the scene (\eg, a person opens a fridge, or washes dishes in a sink), which is typical in applications such as robotics, navigation or surveillance, \etc, we can leverage the interaction between humans and objects to help solve the co-segmentation problem. In essence, as a person interacts with an object in the scene, he/she provides an implicit cue that allows to identify the object's spatial extend (\eg, its segmentation mask) as well as the functional or affordances properties of the object (\ie, the object regions that an human touches in order to use it).

Therefore, in this work, we propose a \emph{human centred co-segmentation} method whereby the common objects are those often used by the observed humans in the scene and sharing the similar human-object interactions such as the pots and the sinks in Fig.~\ref{fig:task}-(c). We show that leveraging this information improves the results considerably compared to previous co-segmentation approaches.

The main challenge of a human centred object co-segmentation is to modeling the rich relations between objects as well as objects and humans in the image set. To achieve this, we first generate a set of object proposals as foreground candidates from the images. In order to discover the rich internal structure of these proposals reflecting their human-object interactions and visual similarities, we then leverage the power and flexibility of the fully connected conditional random field (CRF)~\cite{Philipp_NIPS_2011} in a unsupervised setting and propose a \emph{fully connected CRF auto-encoder}.

Our model uses the fully connected CRF to encode rich features to detect similar objects from the whole dataset. The similarity depends not only on object visual features but also a novel human-object interaction representation. We propose an efficient learning and inference algorithm to allow the full connectivity of the CRF with the auto-encoder, that establishes pairwise similarities on all pairs of the proposals in the dataset. As a result, the model selects the object proposals which have the most human interactions and are most similar to other objects in the dataset as the foregrounds.  Moreover, the auto-encoder allows to learn the parameters from the data itself rather than supervised learning~\cite{Jiang_CVPR_2013,Jiang_RSS_2013} or manually assigned parameters~\cite{Fu_CVPR_2015,Fu_CVPR_2014,Wu_RSS_2014} as done in conventional CRF.

In the experiments, we show that our human centred object co-segmentation approach improves on the state-of-the-art co-segmentation algorithms on two human activity key frame Kinect datasets and a musical instrument RGB image dataset. To further show the generalization ability of the model, we also show a very encouraging co-segmentation result on a dataset combining the images without humans from the challenging Microsoft COCO dataset and the images with tracked humans.

In summary, the main contributions of this work are:
\vspace{-0.05in}
\begin{itemize}
\item We are the first to demonstrate that modeling human is useful to mining common objects more accurately in the unsupervised co-segmentation task.
\item We propose an unsupervised fully connected CRF auto-encoder, and an efficient learning and inference approach to modeling rich relations between objects and humans.
\item We show the leading performance of our human centred object co-segmentation in the extensive experiments on four datasets.
\end{itemize}
\vspace{-0.1in}


\vspace{-0.05in}
\section{Related Work}\label{sec:re}
\vspace{-0.05in}
\header{Co-segmentation.} Many efforts have been made on co-segmenting multiple images~\cite{Brostow_ECCV_2008,Joulin_CVPR_2010,Manfredi_ICIP_2014,Baraldi_ICME_2015,Lucchi_TMI_2015}. The early works used histogram matching~\cite{Rother_CVPR_2006}, scribble guidance~\cite{Batra_IJCV_2011}, or discriminative clustering~\cite{Hochbaum_ICCV_2009} based on low-level descriptors to extract common foreground pixels. A mid-level representation using
``objectness'' was considered in ~\cite{Vicente_CVPR_2011,Meng_TMM_2012} to extract similarly looking foreground objects rather than just common regions. Recently, Fu \etal~\cite{Fu_CVPR_2015} proposed an object-based co-segmentation from RGB-D images using the CRF models with mutex constraints. Our work also generates the object proposals using ``objectness'' and selects the foregrounds from the candidates. Differently, we are the first one to consider the human interaction to extract the foreground objects.

Early works on co-segmentation only considered two foreground and background classes. Recently, there are many co-segmentation methods which are able to handle multiple foreground objects. Kim \etal~\cite{Kim_ICCV_2011} proposed an anisotropic diffusion method by maximizing the overall temperature of image sites associated with a heat diffusion process. Joulin \etal~\cite{Joulin_CVPR_2012} presented an effective energy-based method that combines a spectral-clustering term with a discriminative term, and an efficient expectation-minimization algorithm to optimize the function. Lopamudra \etal~\cite{Lopamudra_ECCV_2012} proposed a method by analyzing the subspace structure of related images. In~\cite{Chiu_CVPR_2013,Fu_CVPR_2014}, they presented a video co-segmentation method that extracts multiple foreground objects in a video set. Our work is also able to extract multiple foreground objects by formulating a fully connected CRF auto-encoder, which learns rich information from both objects and humans to extract the foregrounds.

\header{Human-Object Interactions.} Modeling human-object interactions or object affordances play an important role in recognizing both objects and human actions in previous works. The mutual context of objects and human poses were modeled in~\cite{Yao_TPAMI_2012} to recognize human-object interactions that improve both human pose estimation and object detection. In~\cite{Desai_CVPRW_2010,Delaitre_NIPS_2011,Escorcia_ICCVW_2013}, human-object interactions in still images or videos were modeled to improve action recognition. Wei \etal~\cite{Wei_ICCV_2013} modeled 4D human-object interactions to improve event and object recognition. In~\cite{Jiang_CVPR_2013,Jiang_RSS_2013}, hallucinated humans were added into the environments and human-object relations are modeled using the latent CRF to improve scene labeling.
In this work, we show that human-object interactions provide an important evidence in the unsupervised mining of common objects from images. We also present a novel human-object interaction representation .

\header{Learning Models.} Our learning and inference model is extended from the work of CRF auto-encoder~\cite{Ammar_NIPS_2014}, which focuses on part-of-speech induction problems using a linear chain sequential latent structure with first-order Markov properties. However, the training and inference become impractical when using fully connected CRF with the auto-encoder. Thus, we introduce an efficient mean field approximation to make the computation still feasible. Our work is also close to the fully connected CRF~\cite{Philipp_NIPS_2011} for the supervised semantic image segmentation, which also uses the mean field approximation to achieve an efficient inference. In contrast with this approach,, we use the mean field approximation to fast compute the gradients of two partition functions, which are exponential growing with object classes without the approximation in the CRF auto-encoder.


  \vspace{-0.05in}
\section{Problem Formulation}  
\vspace{-0.05in}
Our goal is to segment out common foreground objects from a given set of images. Similar to most co-segmentation approaches, we first generate a set of object proposals $X=\{x_i\}_{i=1,2,\cdots,N}$ as foreground candidates from each image, where $N$ is the total number of object proposals in the given set of images. Here we use selective search~\cite{Uijlings_IJCV_2013}, which merges superpixels to generate proposals based on the hierarchical segmentation. This approach has been broadly used as the proposal method of choice by many state-of-the-art object detectors. 

Let us assume there are $K$ objects in the images. We then formulate the object co-segmentation as a probabilistic inference problem. We denote the object cluster assignment of $x_i$ as $y_i\in{\mathcal{Y}=[1,2,\cdots,K]}$ and $Y=\{y_i\}_{i=1,2,\cdots,N}\in{\mathcal{Y}^N}$. We want to infer the cluster assignment of a proposal using other proposals with the probability $p(y_i|X)$. We select the object proposals with the highest inference probabilities as the foregrounds, since they have the most human interactions and similar objects as determined by our probabilistic model.

\begin{figure}[t]
  \begin{center}
  \includegraphics[width=0.7\linewidth]{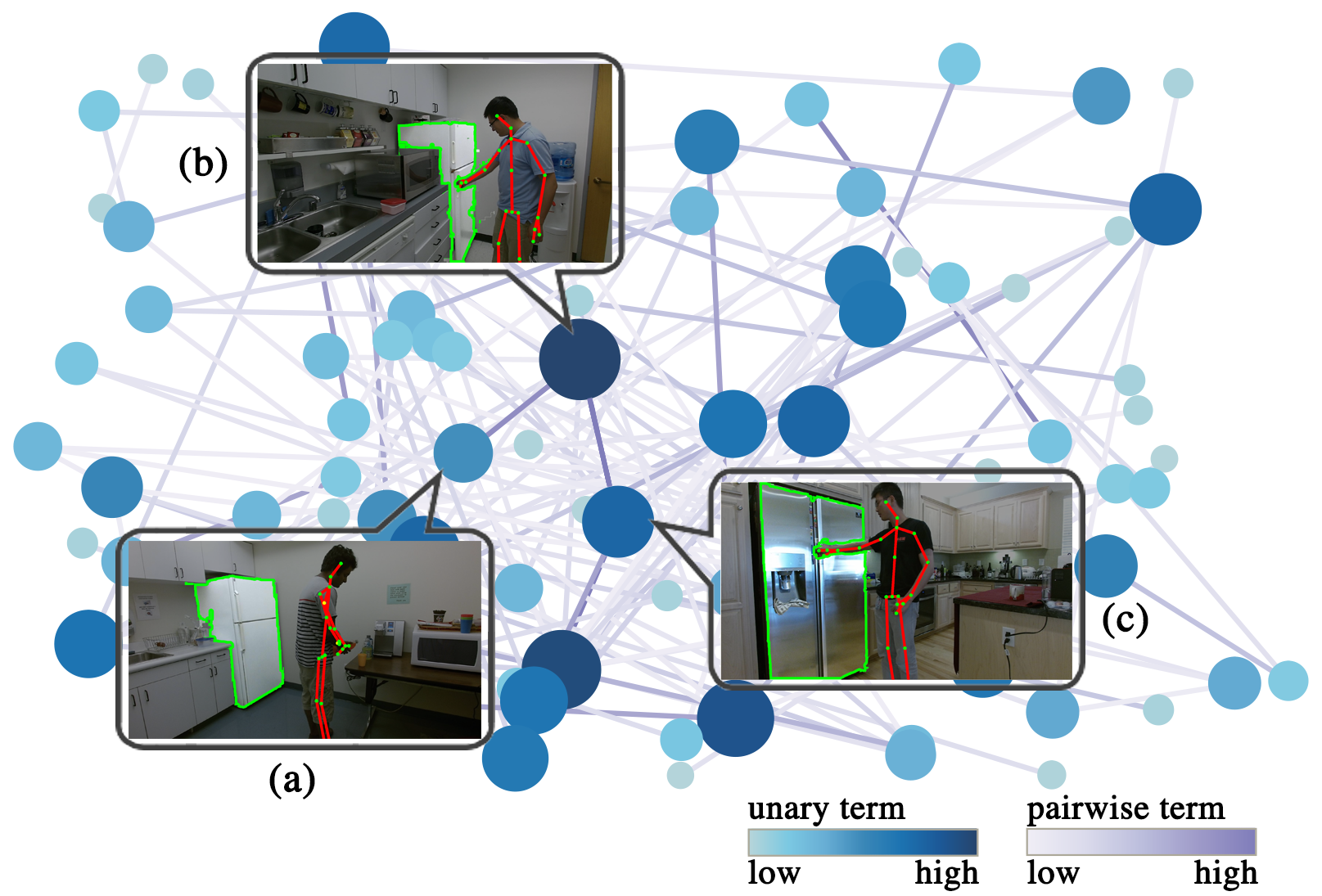}
  \vspace{-0.1in}
  \caption{Learned CRF graph from the data. The nodes are unary terms of object proposals encoding object appearance and human-object interaction features. The edges are pairwise terms encoding similarities on object appearance and human-object interactions. In the example, the fridges in (a) and (b) are visually similar, and the fridges in (b) and (c) have the similar human-object interactions. These similar objects with more human interactions are more likely to be segmented out in our approach, since they have higher unary terms and pairwise terms in the CRF graph. For a good visualization, we only plot the terms with respect to the most likely object cluster for each object proposal and the edges below a threshold are omitted.}
   \label{fig:model_exp}
 \end{center}
\vspace{-0.3in}
\end{figure}
  \vspace{-0.05in}
\section{Model Representation}\label{sec:model}
  \vspace{-0.05in}
We propose a fully connected CRF auto-encoder to discover the rich internal structure of the object proposals reflecting their human-object interactions and visual similarities. Unlike previous works  relying mainly on the visual similarities between objects, we also encourage the objects with more human interactions and having more similar interactions with other objects to be segmented out. We plot a learned CRF graph from a set of images in Fig.~\ref{fig:model_exp}. In the example, the fridges in (a) and (b) are visually similar, and the fridges in (b) and (c) are similar to each other on human-object interactions even though they look different. These similar objects with more human interactions are more likely to be segmented out in our approach, since they have higher terms in the CRF graph.

\begin{figure}[t]
  \begin{center}
  \includegraphics[height=1.4in]{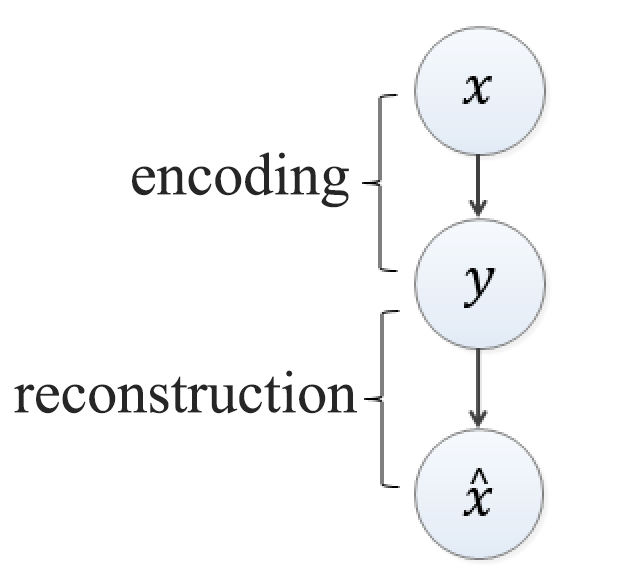}\hspace{0.2in}
  \includegraphics[height=1.4in]{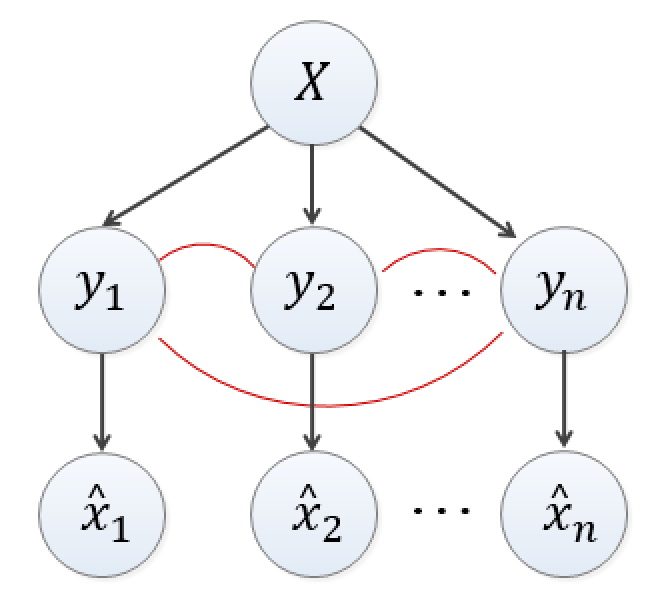}
\vspace{-0.1in}
  \caption{Graphic model of our fully connected CRF auto-encoder.}
   \label{fig:fcrfa}
 \end{center}
\vspace{-0.3in}
\end{figure}

The fully connected CRF auto-encoder consists of two parts (The graphic model is shown in Fig.~\ref{fig:fcrfa}). The encoding part is modeled as a fully connected CRF, which encodes the observations $x_i$ of the object proposal into object cluster hidden nodes $y_i$. The reconstruction part reconstructs the hidden nodes by generating a copy of the observation itself $\hat{x}_i$, which considers that a good hidden structure should permit reconstruction of the data with high probability~\cite{Ammar_NIPS_2014}, 

  \vspace{-0.05in}
\subsection{Fully Connected CRF Encoding}
  \vspace{-0.05in}
We first introduce how the fully connected CRF encodes the observations of the object proposals. In CRF, the conditional distribution is given by $P_{\lambda}(Y|X) = \frac{1}{Z}\exp\{\Phi_\lambda(X,Y)\}$, where $Z=\sum_{Y'\in{\mathcal{Y}^N}}\exp\{\Phi_\lambda(X,Y')\}$ is a partition function and $\Phi_{\lambda}(.)$ is called the energy function defined as follows:
 \begin{equation}\label{eqn:ccrf}
\vspace{-0.1in}
\Phi_{\lambda}(X,Y) = \overbrace{\sum_{i}\lambda^{(u)}(y_i)^{\top}x_i}^{\rm unary\ terms}+ \overbrace{\sum_{i<j}\lambda^{(p)}(y_i,y_j)S(x_i,x_j)}^{\rm pairwise\ terms},
\end{equation}
where $\lambda^{(u)}(y_i)^{\top}x_i$ is the unary term that encodes object visual appearance features and human-object interaction features:
\begin{equation}\label{eqn:unary}
\vspace{-0.1in}
\lambda^{(u)}(y_i)^{\top}x_i = \overbrace{\lambda^{(uo)}(y_i)^{\top}f_i}^{\rm object\ appearance} + \overbrace{\lambda^{(uh)}(y_i)^{\top}h_i}^{\rm human-object\ interaction}.
\end{equation}

\header{Object Appearance Modeling.}
In Eq.~(\ref{eqn:unary}), $\lambda^{(uo)}(y_i)^{\top}f_i$ encodes the object visual feature vector $f_i$ by the linear weights $\lambda^{(uo)}(k)$ of each cluster $k$. It encourages an object to give larger value $\lambda^{(uo)}(k)^{\top}f_i$ if it belongs to the object cluster $k$. We use rich kernel descriptors by kernel principal
component analysis~\cite{Bo_NIPS_2010} on the input images:  gradient, color, local binary pattern for the RGB image, depth gradient of depth image and spin, surface normals of point cloud for the Kinect data, which have been proven to be useful features for scene labeling~\cite{Ren_CVPR_2012,Wu_RSS_2014}.

\header{Human-Object Interaction Modeling.}
In Eq.~(\ref{eqn:unary}), $\lambda^{(uh)}(y_i)^{\top}h_i$ encodes the human-object interaction feature vector $h_i$ by the linear weights $\lambda^{(uh)}(k)$ of each cluster $k$.

To capture the interactions between objects and humans, we propose a novel feature to represent physical human-object interactions such as sitting on the chair, opening the fridge, using their spatial relationships. This feature helps detect those objects used by humans in the scene.

\begin{figure*}[t]
  \begin{center}
  \subfigure[]{
  \includegraphics[height=2.cm]{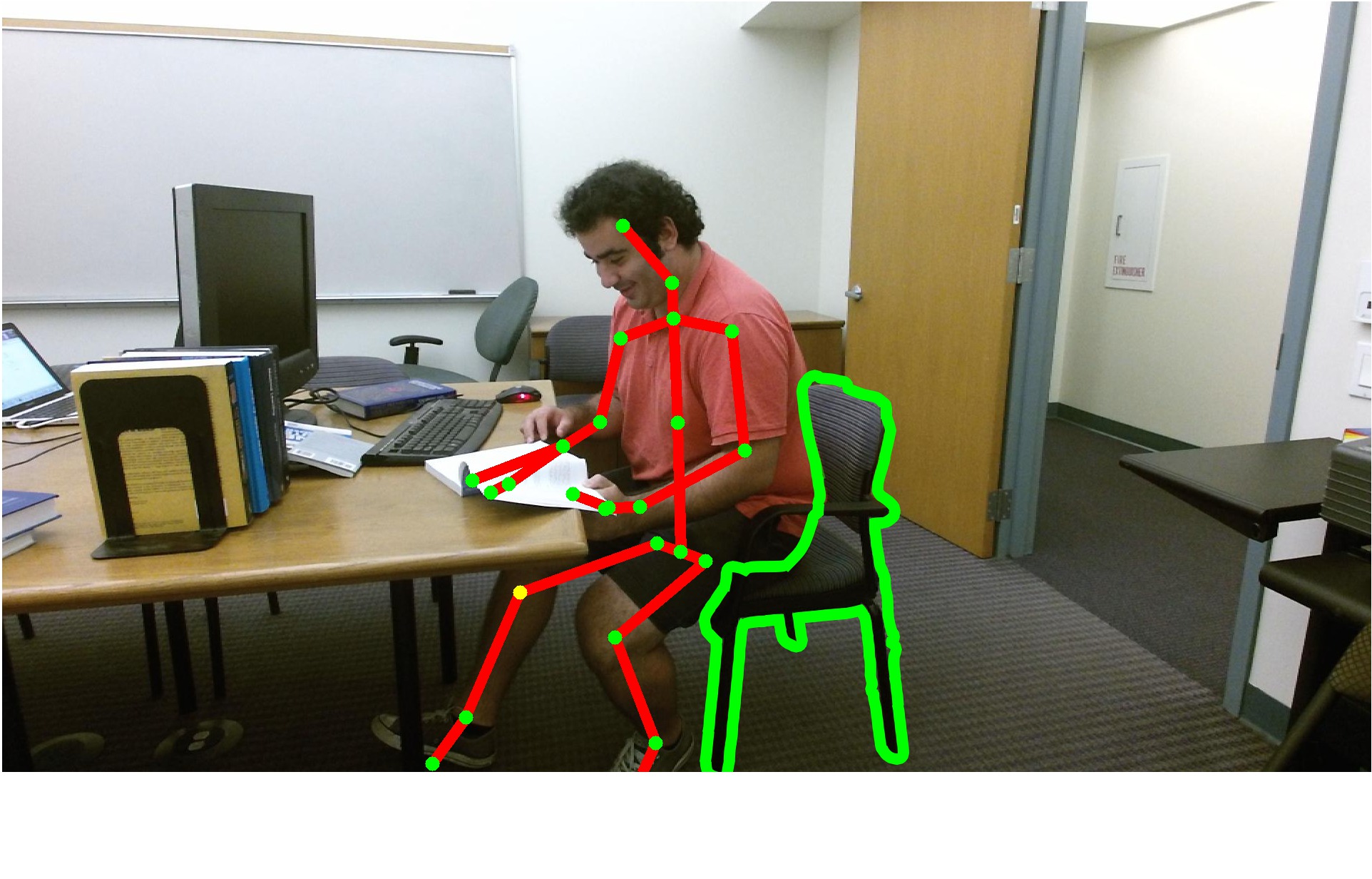}
  }\hspace{-0.05in}
  \subfigure[]{
  \includegraphics[height=2.cm]{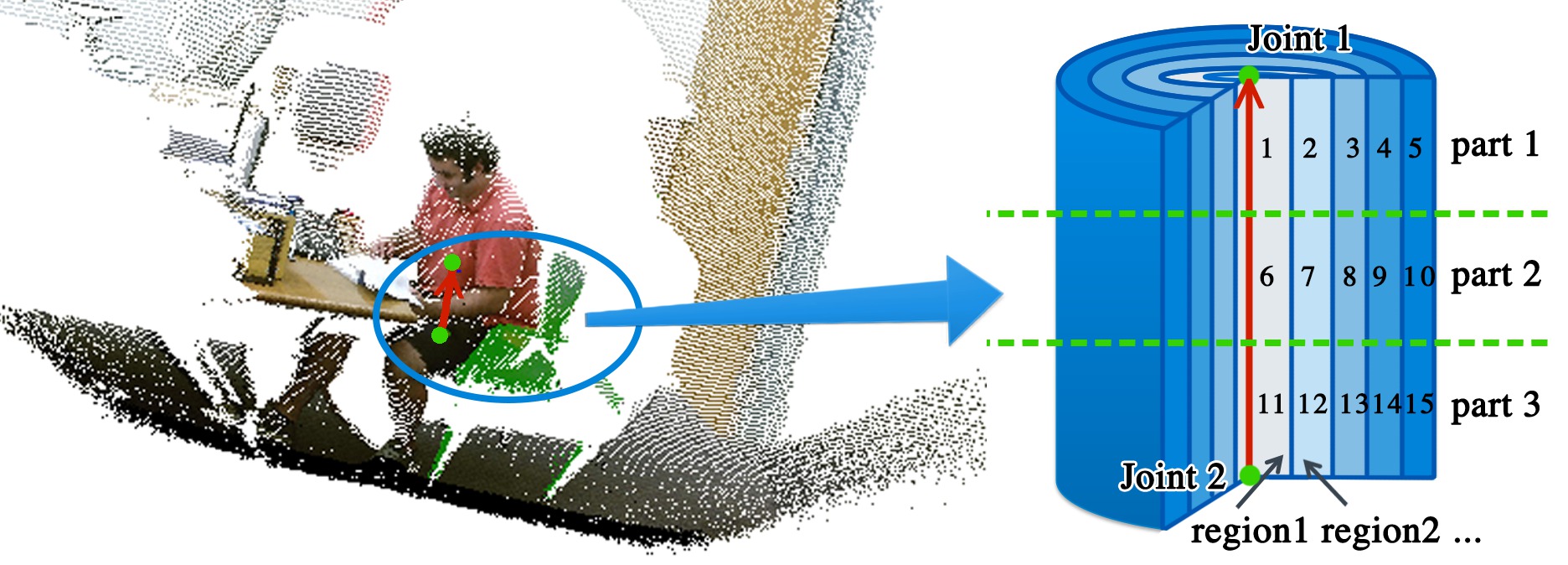}\label{fig:fea_3d}
  }\hspace{-0.1in}
   \subfigure[]{
  \includegraphics[height=2.3cm]{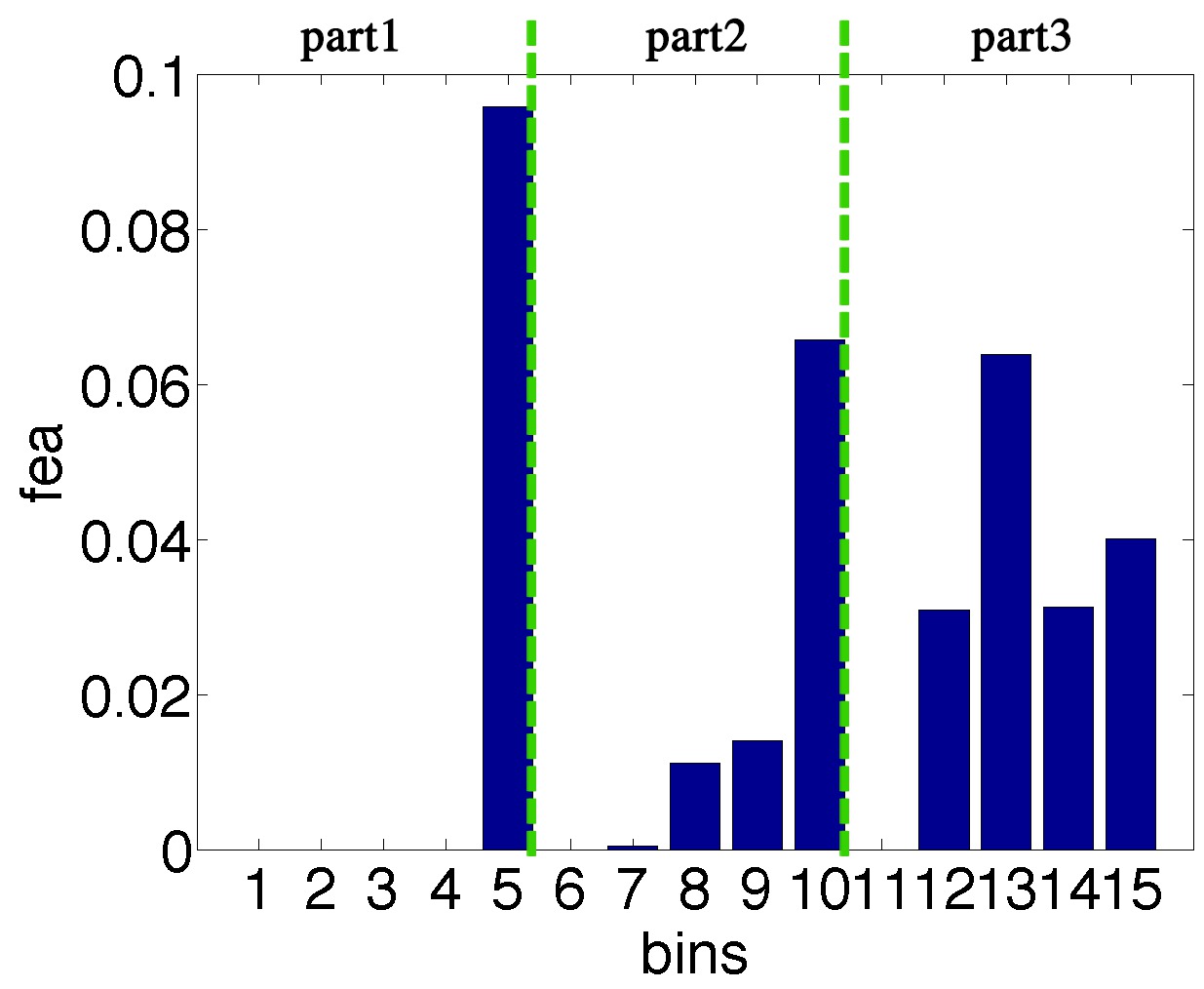}\label{fig:fea_hist}
  }
\vspace{-0.1in}
  \caption{Our human-object interaction feature for RGB-D data. In (a), we are given the joints (green dots) of the tracked human, and a object proposal region (green mask). In (b), we divide the cylinder surrounding each body part vector (red line, \emph{spine-base} to \emph{spine-mid} in the example) into $15$ bins by segmenting the body part vertically into $3$ parts and the circle surrounding the body part into $5$ regions. In (c), we compute the histogram of the points in these $15$ bins and normalize it by the total number of the total points in the object proposal.}
   \label{fig:hoif}
 \end{center}
\vspace{-0.35in}
\end{figure*}

We illustrate the feature representation in Fig.~\ref{fig:hoif} for RGB-D data. In detail, we convert the depth image into the real-world 3D point cloud and are also given the 3D coordinate of each joint of a tracked human. Each human body part is represented as a vector starting from a joint to its neighboring joint (Fig.~\ref{fig:hoif}-(a)(b)). Then we consider a cylinder with the body part as the axis and divide the cylinder into $15$ bins by segmenting the body part vertically into $3$ parts and the circle surrounding the body part into $5$ regions evenly\footnote{To avoid the affect of points of the human part, we do not consider the innermost circle.} (Fig.~\ref{fig:hoif}-(b)). Given the point cloud in an object proposal region, we calculate the histogram of the points in these $15$ bins and normalize it by the number of the total points in the object proposal as the final feature $h_i$ (Fig.~\ref{fig:hoif}-(c)). For multiple people in the scene, we compute the max of the histograms of all humans. 

For RGB only data, we assume a 2D bounding box of human is given by a person detector. We divide the bounding box evenly into $6\times 6$ bins and compute the normalized histogram of pixels within the bins. 
 
The feature captures the distributions of the object points relative to the human body. It can represent the relative position, size between the humans and objects as well as the human poses and object shapes especially in 3D space. For example in Fig.~\ref{fig:hoif}-(c), we plot the histogram feature of the body part \emph{spine-base} to \emph{spine-mid} relative to the \emph{chair}. We can see that the distribution reflects the shape of the chair in the following way: the points of the chair-back lie in the distant bins of the upper part (mostly in bin $5$ among bin $1$-$5$), and from the chair-back to the chair-base, points become more evenly distributed from the middle part (bin $6$-$10$) to the lower part (bin $11$-$15$).

Since our feature vector $h_i$ has larger values for more human-object interactions, we constrain the weights of the human-object interaction features $\lambda^{(uh)}(k) \geq 0$ to encourage more interactions.

\header{Pairwise Similarity Modeling.}
In Eq.~(\ref{eqn:ccrf}), $\lambda^{(p)}(y_i,y_j)S(x_i,x_j)$ is the pairwise term that encodes similarities on object visual appearance and human-object interactions of each object pair:
\vspace{-0.05in}
\begin{equation}
\lambda^{(p)}(y_i,y_j)S(x_i,x_j) = \overbrace{\lambda^{(po)}(y_i,y_j)S(f_i,f_j)}^{\rm object\ similarities} + \overbrace{\lambda^{(ph)}(y_i,y_j)S(h_i,h_j)}^{\rm interaction\ similarities} ,
\end{equation}
where $S(a,b)=\exp\{\frac{-\|a-b\|_2^2}{\delta}\}$ is the similarity function between features\footnote{In practice, we use the augmented features $f'_i =[f_i,-1], h'_i = [h_i,-1],S'(x_i,x_j) = [S(x_i,x_j),-1]$ to also learn the bias of the features. So $\lambda^{(uo)}(k) = [\omega^{(uo)},b^{(uo)}],\lambda^{(uh)}(k)=[\omega^{(uh)},b^{(uh)}],\lambda^{(p)}(k,k)= [\omega^{(p)},b^{(p)}]$ also add another dimension, where $\omega$ are the importance weights and $b>0$ is the bias.} . Note that we consider same class to be similar to each other, so we constrain $\lambda^{(p)}(k,k)>0$ and $\lambda^{p}(k,l) = 0, k\neq l$.

\vspace{-0.05in}
\subsection{Reconstruction}
\vspace{-0.05in}
To independently generate $\hat{x}$ given $y$, we define the reconstruction model as a multivariate normal distribution:
\vspace{-0.05in}
\begin{equation}\label{eqn:rec}
\vspace{-0.05in}
P_\theta(\hat{x}|y) = N(\hat{x}|\theta^{(\mu)}(y),\theta^{(\Sigma)}(y)),
\end{equation}
where $\theta^{(\mu)}(k),\theta^{(\Sigma)}(k)$ are the mean and covariance of the normal distributions of each class $k$.

\vspace{-0.05in}
\section{Model Learning and Inference}
\vspace{-0.05in}
In this section, we introduce how we learn the parameters in our fully connected CRF autoencoder from the data and the inference.

Eq.~\ref{eqn:obj} gives the parametric model for the observations $X$:
\vspace{-0.1in}
\begin{equation}\label{eqn:obj}
\begin{split}
&P_{\lambda,\theta}(\hat{X}|X) = \sum_{Y\in{\mathcal{Y}^N}}\overbrace{P_{\lambda}(Y|X)}^{\rm encoding}\overbrace{P_{\theta}(\hat{X}|Y)}^{\rm reconstruction}\\
                                = &\sum_{Y\in{\mathcal{Y}^N}}\frac{\exp\{\Phi_\lambda(X,Y)\}}{\sum_{Y'\in{\mathcal{Y}^N}}\exp\{\Phi_\lambda(X,Y')\}}\prod_{i}P_\theta(\hat{x}_i|y_i)\\
                                = &\frac{\sum_{Y\in{\mathcal{Y}^N}}\exp\{\Phi_\lambda(X,Y)+\log\sum_{i}P_\theta(\hat{x}_i|y_i)\}}{\sum_{Y'\in{\mathcal{Y}^N}}\exp\{\Phi_\lambda(X,Y')\}},\\                     
\end{split}
\end{equation}
where $\lambda,\theta$ are the parameters of the encoding parts $P_{\lambda}(Y|X)$ and the reconstruction parts  $P_{\theta}(\hat{X}|Y)$. 


\vspace{-0.05in}
\subsection{Efficient Learning and Inference} \label{sec:learn}
\vspace{-0.05in}
We maximize the regularized conditional log likelihood of $P_{\lambda,\theta}(\hat{X}|X)$ over all the object proposals to learn the parameters $\lambda,\theta$:
\vspace{-0.1in}
\begin{equation}\label{eqn:lobj}
\log L(\lambda,\theta) = R_1(\lambda)+R_2(\theta) + \log\sum_{Y}P_{\lambda}(Y|X)P_{\theta}(\hat{X}|Y),
\end{equation}
where $R_1(\lambda),R_2(\theta)$ are the regularizers. Note that the space of $Y$ is $[1,2,\cdots, N]^K$, which is exponential to object cluster number.  Maximizing the above function requires computing gradients, while the summation over $Y,Y'$ in the space  is intractable.

\header{Mean Field Approximation.}
In order to make the learning and inference efficient, we use a mean field approximation that has been widely used in the complex graph inference~\cite{Koller_PGM_2009,Philipp_NIPS_2011}. To use mean field approximation in our model, we first introduce two probabilities of $Y$:
\vspace{-0.05in}
\begin{equation}\label{eqn:prob}
P_{\lambda,\theta}(Y) = \frac{1}{Z}\exp\{\Phi_\lambda(X,Y) + \log P_{\theta}(\hat{X}|Y)\}, \ \ 
P'_{\lambda}(Y) = \frac{1}{Z'}\exp\{\Phi_\lambda(X,Y)\},
\end{equation}
where 
\vspace{-0.05in}
\begin{equation}
Z = \sum_{Y\in{\mathcal{Y}^n}}\exp\{\Phi_\lambda(X,Y) + \log P_{\theta}(\hat{X}|Y)\}, \ \ 
Z' = \sum_{Y\in{\mathcal{Y}^n}}\exp\{\Phi_\lambda(X,Y)\},
\end{equation}
are the partition functions summing over $Y$ to make two probabilities valid. Then Eq.(\ref{eqn:lobj}) can be rewritten as:
\begin{equation}\label{eqn:slobj}
\log L(\lambda,\theta)  = R_1(\lambda) + R_2(\theta) + \log Z-\log Z'.
\end{equation}

Instead of directly computing $P_{\lambda,\theta}(Y),P'_{\lambda}(Y)$, we use the mean field approximation to compute the distributions $Q(Y),Q'(Y)$ that minimize the KL-divergence $D(Q||P_{\lambda,\theta}),D(Q'||P'_{\lambda})$. Then all distributions $Q,Q'$ can be expressed as a product of independent marginals, $Q(Y) = \prod_iQ_i(y_i), Q'(Y) = \prod_iQ'_i(y_i)$~\cite{Koller_PGM_2009}. As a result, the gradients of $\log Z$ and $\log Z'$ can be easily computed approximately (see Eq.~(\ref{eqn:grad})).

By minimizing the KL-divergence, $Q_i(Y),Q'_i(Y)$ has the following iterative update equations:
\vspace{-0.05in}
\begin{equation}
\begin{split}
Q_i(y_i=k) =& \frac{1}{Z_i}\exp\{\lambda^{(u)}(k)^{\top}x_i+\lambda^{(p)}(k,k)\sum_{j\neq i}S(x_i,x_j)Q_j(k) \\
 &+\log N(\hat{x}_i|\theta^{(\mu)}(k),\theta^{(\Sigma)}(k))\},\\
Q'_i(y_i=k) =& \frac{1}{Z'_i}\exp\{\lambda^{(u)}(k)^{\top}x_i+
\lambda^{(p)}(k,k)\sum_{j\neq i}S(x_i,x_j)Q'_j(k)\}.
\end{split}
\end{equation}
The detailed derivation is in the supplementary material. Following the above equations, the computation of $Q,Q'$ can be done using an iterative update as shown in Algorithm~\ref{alg:mf}.

In each update, it costs $O(N^2\times K)$. It also can be used a similar message passing by the high-dimensional filtering as in~\cite{Philipp_NIPS_2011} to speed up the update, which is not the focus of the paper. The update was mostly converged within $10$ rounds in our experiments (see Fig.~\ref{fig:cv}).

\begin{algorithm*}[t]
\caption{Mean field to compute $Q$ and $Q'$.}
\begin{algorithmic}
\STATE Initialize $Q$ and $Q'$:
\STATE \ \ \ $Q_i(k)=\frac{1}{Z_i}\exp\{\lambda^{(u)}(k)^{\top}x_i+\log N(\hat{x}_i|\theta^{(\mu)}(k),\theta^{(\Sigma)}(k))\}$,
\STATE \ \ \ $Q'_i(k)=\frac{1}{Z'_i}\exp\{\lambda^{(u)}(k)^{\top}x_i\}$,
\WHILE{not converged}
\STATE \ \ \ $\hat{Q}_i(k) = \lambda^{(p)}(k,k)\sum_{j\neq i}S(x_i,x_j)Q_j(k)$,
\STATE \ \ \ $Q_i(k) = \exp\{\lambda^{(u)}(k)^{\top}x_i+\hat{Q}_i(k)+\log N(\hat{x}_i|\theta^{(\mu)}(k),\theta^{(\Sigma)}(k))\}$,
\STATE \ \ \ $\hat{Q'}_i(k) = \lambda^{(p)}(k,k)\sum_{j\neq i}S(x_i,x_j)Q'_j(k)$,
\STATE \ \ \ $Q'_i(k) = \exp\{\lambda^{(u)}(k)^{\top}x_i+\hat{Q'}_i(k)\}$,
\STATE \ \ \ normalize $Q_i(k),Q'_i(k)$.
\ENDWHILE
\end{algorithmic}  \label{alg:mf}
\end{algorithm*}
\header{Learning.} We iteratively learn $\lambda$ and $\theta$ by maximizing the log likelihood with respect to each type of parameters. We update $\lambda$ using the Adagrad approach~\cite{Duchi_JMLR_2011} which computes the gradients and update $\theta$ using EM~\cite{Dempster_EM_1977}. Using the mean field approximation described above, we can easily compute the gradients of $\log Z,\log Z'$ in Eq.(\ref{eqn:slobj}), leading to a simple approximation of the gradients:
\begin{equation}\label{eqn:grad}
\begin{split}
&\frac{\partial (\log Z-\log Z')}{\partial \lambda^{(u)}(k)}
= \sum_ix_i(Q_i(k)-Q'_i(k)),\\
&\frac{\partial (\log Z-\log Z')}{\partial \lambda^{(p)}(k,k)}
= \sum_{i\neq j}S(x_i,x_j)(Q_i(k)Q_j(k)-Q'_i(k)Q'_j(k)).
\end{split}
\end{equation}
The detailed derivation is in the supplementary material. 

\header{Inference.} After learning the parameters $\lambda,\theta$, we can infer the posterior, conditioning on both observations and reconstructions, $\hat{Y}=\arg\max_{Y}P_{\lambda,\theta}(Y|X,\hat{X})$. This is proportional to $P_{\lambda,\theta}(Y)$ in Eq.(\ref{eqn:prob}), which can be efficiently computed using the approximation probability $Q(Y) = \prod_iQ_i(y_i)$. So the probability of each object proposal for each class $p_{\lambda,\theta}(y_i|X,\hat{X})\propto Q_i(y_i)$, which we use as the confidence score to extract the foreground objects of each image.


\if -
\begin{table}[t]
\setlength{\tabcolsep}{10pt}
\begin{center}
\caption{Co-Segmentation result on PPMI dataset (\%).}\label{tb:reppmi}
\begin{tabular*}{\linewidth}{@{\extracolsep{\fill}}c|c c c c}
\hline
class&Ours&\cite{Joulin_CVPR_2012}&\cite{Fu_CVPR_2014}\\
\hline\hline
table&50.0&34.7&21.6\\
chair&36.4&17.2&15.7\\
\hline
fridge&44.7&29.9&25.4\\
microwave&10.5&5.5&24.5\\
\hline
pod&17.7&5.3&\\
sink&28.6&17.9&\\
\hline
\end{tabular*}
\end{center}
\vspace{-0.1in}
\end{table}
\fi

\vspace{-0.05in}
\section{Experiments}\label{sec:exp}
\vspace{-0.05in}
\subsection{Compared Baselines}
\vspace{-0.05in}
We compared our human centred object co-segmentation approach with two state-of-the-art algorithms: a multi-class image co-segmentation approach by combining spectral- and discriminative-clustering (Joulin \etal 2012~\cite{Joulin_CVPR_2012}) and a CRF based solution with the manually assigned parameters (Fu \etal 2014~\cite{Fu_CVPR_2014}). Fu \etal 2014~\cite{Fu_CVPR_2014} is designed for co-segmenting foregrounds from videos, so we keep their features for the static frame and remove the temporal features. We run their source code on the given RGB images with the default settings. We also evaluate our model using object-only visual features in the experiments.   

\vspace{-0.05in}
\subsection{Evaluations}
\vspace{-0.05in}
We use the same evaluation metrics as in multi-class co-segmentation~\cite{Joulin_CVPR_2012}: the intersection-over-union score defined as $\max_k\frac{1}{|\I|}\sum_{i\in{\I}}\frac{GT_i\cap R_i^k}{GT_i\cup R_i^k}$, where $\I$ is the image dataset, $GT_i$ is the ground-truth region and $R_i^k$ is the region associated with the $k$-th cluster in image $i$. The evaluation validates that the foreground object is indeed rather well represented by one of the object clusters in most test cases, which may be sufficient to act as a filter in most applications such as~\cite{Russell_CVPR_2006}. As discussed in multi-class co-segmentation~\cite{Joulin_CVPR_2012}, we set $K$ to be more than the number of groundtruth foreground object classes plus background class to give better segmentation results. We set $K=4$ for the data with a single foreground and $K=6$ for the data with two foregrounds. We also evaluate how performance varies with $K$ in the experiments.

\vspace{-0.05in}
\subsection{Datasets}
\vspace{-0.05in}
\header{Human Activity Key Frame Kinect Dataset}
We first evaluate on human activity key frames extracted from the activity videos in two datasets: Cornell Activity Dataset-$120$ (CAD-$120$)\footnote{\url{http://pr.cs.cornell.edu/humanactivities/data.php}}~\cite{Koppula_IJRR_2013} recorded by Microsoft Kinect v1, and Watch-n-Patch\footnote{\url{http://watchnpatch.cs.cornell.edu/}}~\cite{Wu_CVPR_2015} recorded by the Microsoft Kinect v2. Each frame in the datasets has a registered RGB and depth image pair as well as tracked human skeletons. 

CAD-$120$ contains activity sequences of ten different high level activities performed by four different subjects. We evaluate the image co-segmentation with the top four existing foreground objects in CAD-$120$: microwave, bowl, box, cup. For each type, we extract the $60$ key frames of the activity videos containing the object. 
We evaluate the foreground regions using the provided groundtruth bounding box.

Watch-n-Patch activity dataset contains human daily activity videos performed by $7$ subjects in $8$ offices and $5$ kitchens with complex backgrounds. In each environment, the activities are recorded in different views. In each video, one person performed a sequence of actions interacting with different types of objects. We evaluate three types of scenes in the dataset, each of which has two foreground objects: table and chair, fridge and microwave, pod and sink. For each scene, we extract the $70$ key frames of the relevant activity video containing the objects. We label the groundtruth foreground pixels for evaluation.

\header{People Playing Musical Instrument Dataset.}
We also evaluate on a RGB dataset to see the performance using RGB only features. We use the People Playing Musical Instrument (PPMI) dataset\footnote{\url{http://ai.stanford.edu/~bangpeng/ppmi.html}}, which was used to evaluate recognizing human-object interactions in~\cite{Yao_TPAMI_2012}. It contains RGB images of people `playing' or `with' different types of musical instruments. Some of the images have multiple people interacting with the instruments. We evaluate three types of instruments, cello, French horn and violin by randomly selecting $80$ images from each class and label the foreground pixels in the image for evaluation.


\header{MS COCO combining with Watch-n-Patch Dataset}
Since image data with humans are not always available, we finally give the co-segmentation results on the images without humans from the challenging Microsoft COCO (MS COCO) dataset~\cite{Lin_COCO_2014} combining with a small portion of images with tracked humans from Watch-n-Patch dataset. The images from MS COCO dataset has more clustered backgrounds from variant sources. We evaluate three classes: chair, fridge and microwave. For each class, we randomly select $50$ images from indoor scenes in MS COCO dataset and $20$ images from Watch-n-Patch dataset, then combine them as the test set. We use the same RGB features and human features as described above.

Note that one challenge for image co-segmentation is the scalability, as the state-of-the-art algorithms rely on heavy computations on relation graphs. Therefore, most evaluation datasets in previous works~\cite{Joulin_CVPR_2010,Vicente_CVPR_2011,Joulin_CVPR_2012,Fu_CVPR_2015} have less than $50$ images per class. In our experiments, the test set has more than $50$ but still less than $100$ images per class.

\begin{table*}[t]
\setlength{\tabcolsep}{4pt}
\begin{center}
\caption{Co-Segmentation results on CAD-120 dataset (\%).}\label{tb:re1}
\vspace{-0.1in}
\begin{tabular}{c|c|c|c|c}
\hline
class&microwave&bowl&box&cup\\
\hline
Joulin \etal 2012~\cite{Joulin_CVPR_2012}&21.6&22.5&19.2&17.7\\
Fu \etal 2014~\cite{Fu_CVPR_2014}&47.5&14.9&\textbf{40.2}&9.3\\
Ours (object-only)&45.1&19.7&32.6&22.8\\
Ours&\textbf{54.3}&\textbf{24.8}&38.2&\textbf{27.9}\\
\hline
\end{tabular}
\end{center}
\vspace{-0.2in}
\end{table*}

\begin{table*}[t]
\setlength{\tabcolsep}{4pt}
\begin{center}
\caption{Co-Segmentation results on Watch-n-Patch dataset (\%).}\label{tb:re2}
\vspace{-0.1in}
\begin{tabular}{c|c c |c c |c c}
\hline
class&table&chair&fridge&microwave&pod&sink\\
\hline
Joulin \etal 2012~\cite{Joulin_CVPR_2012}&34.7&17.2&29.9&5.5&5.3&17.9\\
Fu \etal 2014~\cite{Fu_CVPR_2014}&21.6&15.7&25.4&\textbf{24.5}&21.3&23.6\\
Ours (object-only)&41.9&26.9&33.1&17.5&20.4&23.2\\
Ours&\textbf{50.0}&\textbf{36.4}&\textbf{44.7}&20.5&\textbf{23.7}&\textbf{28.6}\\
\hline
\end{tabular}
\end{center}
\vspace{-0.35in}
\end{table*}

\begin{table*}[t]
\setlength{\tabcolsep}{4pt}
\begin{center}
\caption{Co-Segmentation results on PPMI dataset (\%).}\label{tb:re3}
\vspace{-0.1in}
\begin{tabular}{c|c|c|c}
\hline
class&cello&frenchhorn&violin\\
\hline
Joulin \etal 2012~\cite{Joulin_CVPR_2012}&21.9&20.0&18.1\\
Fu \etal 2014~\cite{Fu_CVPR_2014}&30.1&40.8&26.4\\
Ours (object-only)&34.5&41.0&28.3\\
Ours&\textbf{36.2}&\textbf{49.2}&\textbf{31.5}\\
\hline
\end{tabular}
\end{center}
\vspace{-0.2in}
\end{table*}

\begin{table*}[t]
\setlength{\tabcolsep}{4pt}
\begin{center}
\caption{Co-Segmentation results on MS COCO + Watch-n-Patch dataset (\%).}\label{tb:re4}
\vspace{-0.1in}
\begin{tabular}{c|c|c|c}
\hline
class&chair&fridge&microwave\\
\hline
Joulin \etal 2012~\cite{Joulin_CVPR_2012}&4.2&14.1&10.3\\
Fu \etal 2014~\cite{Fu_CVPR_2014}&6.9&11.4&9.2\\
Ours (object-only)&7.5&10.2&10.5\\
Ours&\textbf{12.5}&\textbf{17.5}&\textbf{15.9}\\
\hline
\end{tabular}
\end{center}
\vspace{-0.35in}
\end{table*}

\vspace{-0.05in}
\subsection{Results}
\vspace{-0.05in}
We give the results in Table~\ref{tb:re1},~\ref{tb:re2},~\ref{tb:re3},~\ref{tb:re4}. We can see that in most cases, our approach performs better than the state-of-the-art image co-segmentation methods. We discuss our results in the light of the following questions.

\begin{figure*}[t]
  \begin{center} 
  \subfigure[table-chair]{
  \begin{minipage}{0.32\linewidth}
  \includegraphics[width=0.47\linewidth]{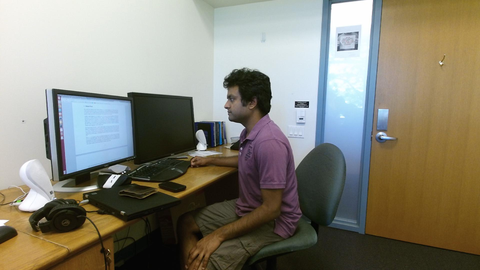}\hspace{-0.02in}
  \includegraphics[width=0.47\linewidth]{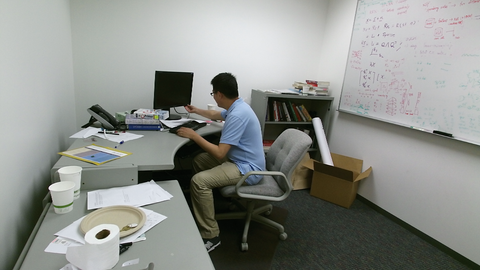}\\
  \includegraphics[width=0.47\linewidth]{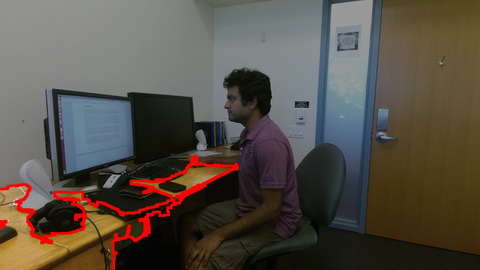}\hspace{-0.02in}
  \includegraphics[width=0.47\linewidth]{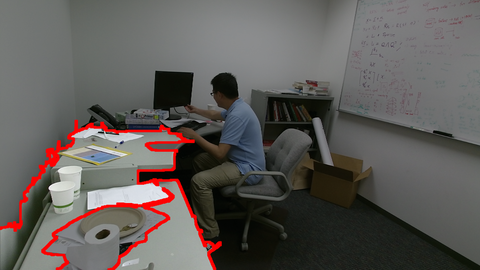}\\
  \includegraphics[width=0.47\linewidth]{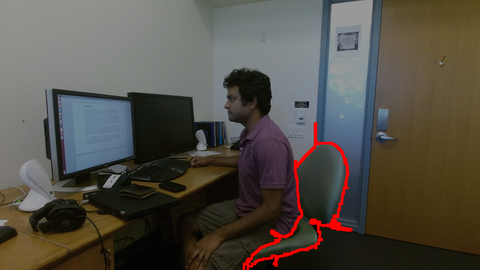}\hspace{-0.02in}
  \includegraphics[width=0.47\linewidth]{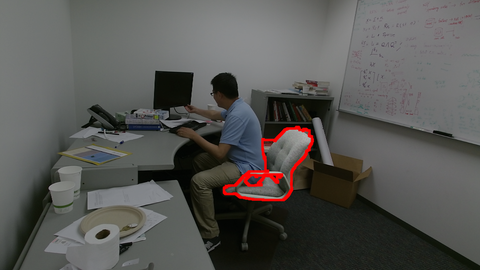}
  \end{minipage}\hspace{-0.15in}
  }\vspace{-0.02in}
  \subfigure[fridge-microwave]{
  \begin{minipage}{0.32\linewidth}
  \includegraphics[width=0.47\linewidth]{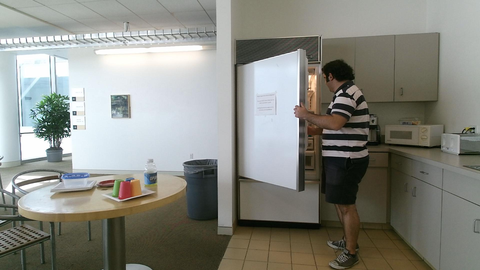}\hspace{-0.02in}
  \includegraphics[width=0.47\linewidth]{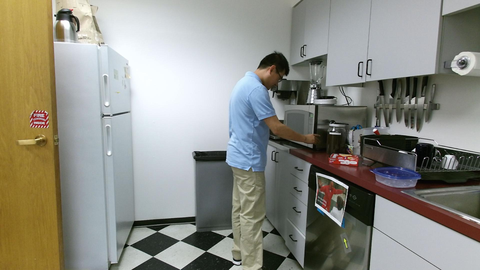}\\
  \includegraphics[width=0.47\linewidth]{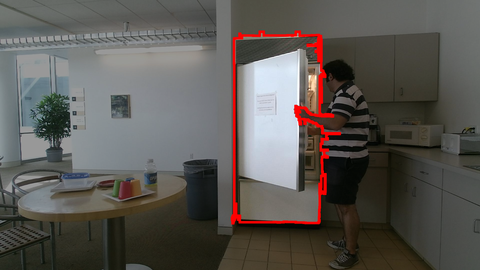}\hspace{-0.02in}
  \includegraphics[width=0.47\linewidth]{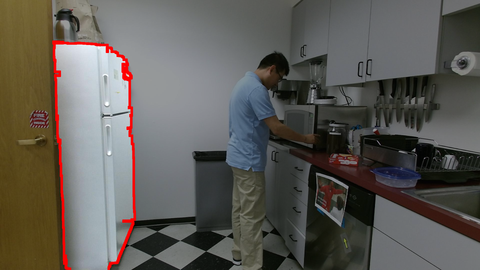}\\
  \includegraphics[width=0.47\linewidth]{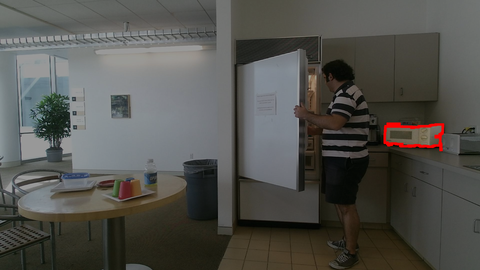}\hspace{-0.02in}
  \includegraphics[width=0.47\linewidth]{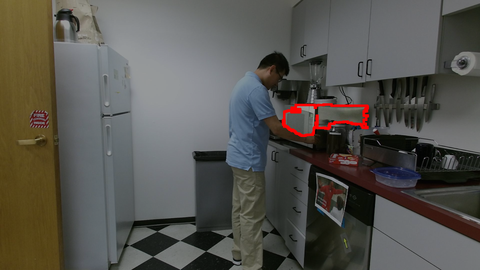}
  \end{minipage}\hspace{-0.15in}\label{fig:ve_fo}
  }\vspace{-0.02in}
  \subfigure[pot-sink]{
  \begin{minipage}{0.32\linewidth}
  \includegraphics[width=0.47\linewidth]{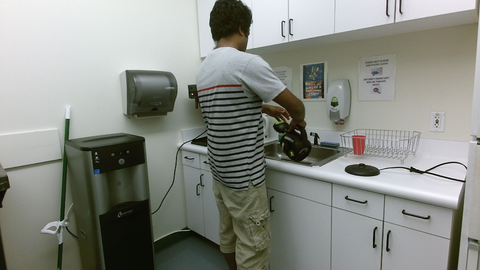}\hspace{-0.02in}
  \includegraphics[width=0.47\linewidth]{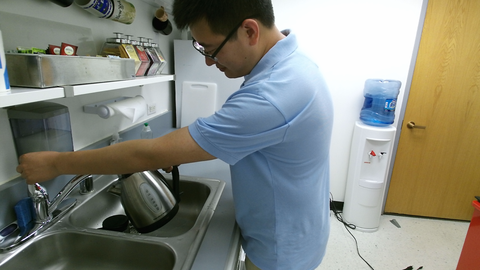}\\
  \includegraphics[width=0.47\linewidth]{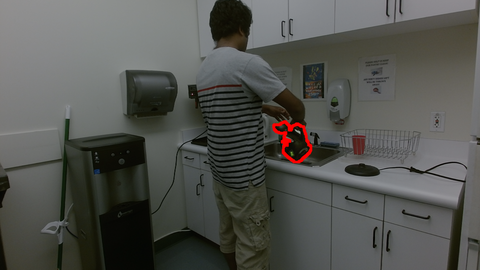}\hspace{-0.02in}
  \includegraphics[width=0.47\linewidth]{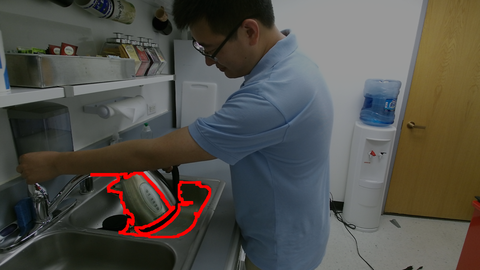}\\
  \includegraphics[width=0.47\linewidth]{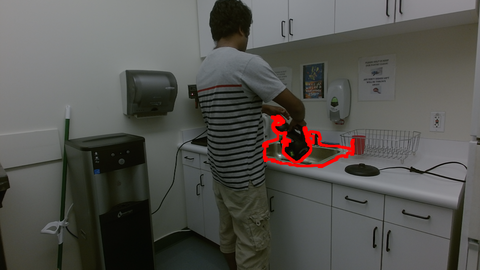}\hspace{-0.02in}
  \includegraphics[width=0.47\linewidth]{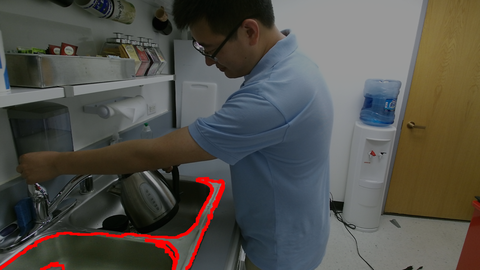}
  \end{minipage}\hspace{-0.1in}
  }\vspace{-0.05in}
  
 \caption{Visual examples of our co-segmentation results on Watch-n-Patch dataset.}
   \label{fig:ve_wnp}
 \end{center}
\end{figure*}

\begin{figure*}[t]
  \begin{center} 
  \hspace{0.05in}
  \subfigure[cello]{
  \begin{minipage}{0.333\linewidth}
  \includegraphics[height = 1.8cm]{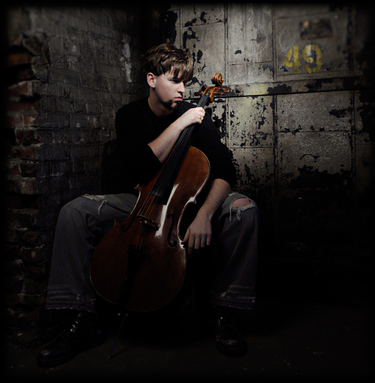}
  \includegraphics[height = 1.8cm]{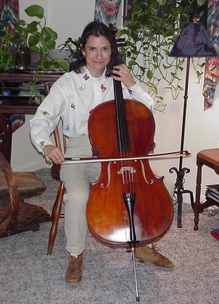}\\
  \includegraphics[height = 1.8cm]{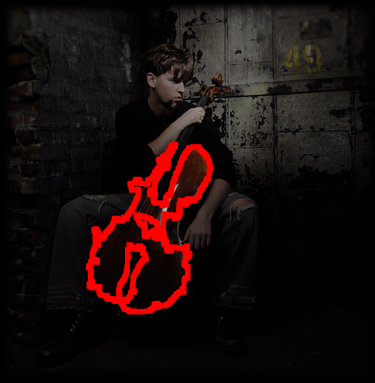}
  \includegraphics[height = 1.8cm]{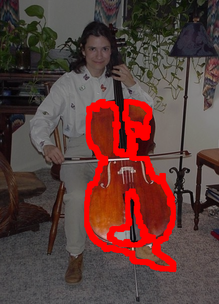}
  \end{minipage}
  }\hspace{-0.35in}
  \subfigure[French horn]{
  \begin{minipage}{0.333\linewidth}
  \includegraphics[height = 1.8cm]{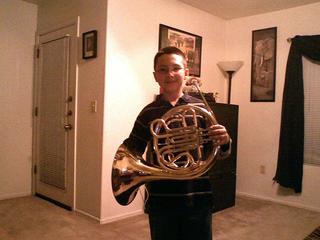}
  \includegraphics[height = 1.8cm]{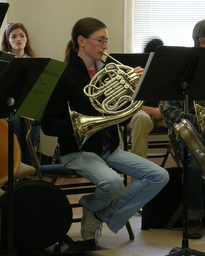}\\
  \includegraphics[height = 1.8cm]{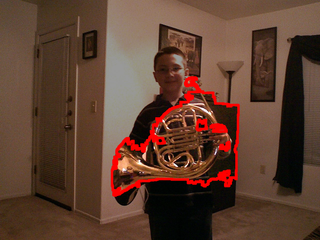}
  \includegraphics[height = 1.8cm]{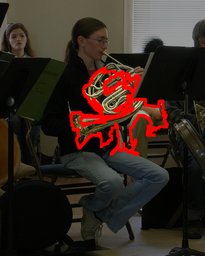}
  \end{minipage}
  }\hspace{-0.05in}
   \subfigure[violin]{
  \begin{minipage}{0.333\linewidth}
  \includegraphics[height = 1.8cm]{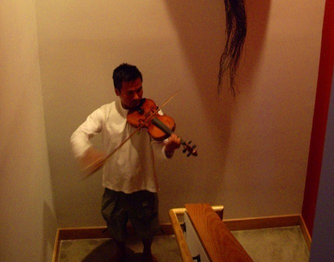}
  \includegraphics[height = 1.8cm]{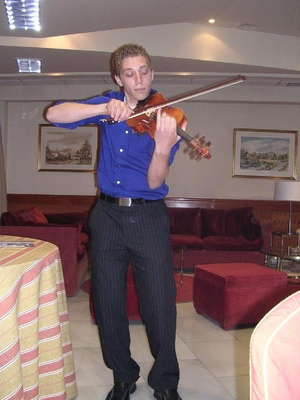}\\
  \includegraphics[height = 1.8cm]{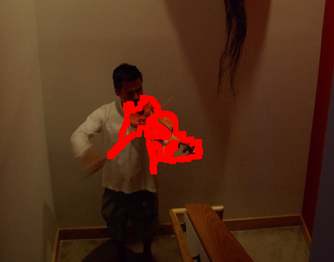}
  \includegraphics[height = 1.8cm]{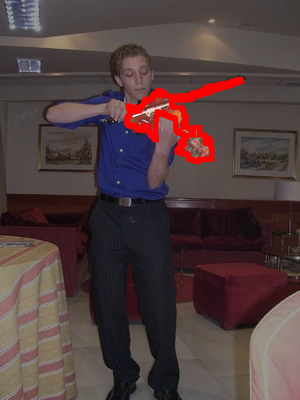}
  \end{minipage}
  }
  \hspace{-0.2in}
  \vspace{-0.05in}
 \caption{Visual examples of our co-segmentation results on PPMI dataset.}
   \label{fig:ve_ppmi}
 \end{center}
 \vspace{-0.1in}
\end{figure*}

\header{Did modeling human-object interaction help?}
In most cases, we can see that our approach to modeling the human-object interaction gives the best result. This is because more human-object interactions give the higher unary term for the interesting objects interacting with the humans, and the similar human-object interactions link the same objects in the CRF graph with larger pairwise terms even though they may not look similar. As a result, we are able to segment out the common object accurately even with view, scale changes, occlusions and in complex backgrounds. We show some example results in Fig.~\ref{fig:ve_wnp} and Fig.~\ref{fig:ve_ppmi}.

\header{How successful is our fully connected CRF auto-encoder?}
From the results, we can see that our fully connected CRF auto-encoder model using the object only features also performs better than other algorithms. This is because our model is able to learn the parameters from the data itself rather than manually assigning parameters of the typical CRF model in Fu \etal 2014~\cite{Fu_CVPR_2014}, then the model is more data dependent and does not require much parameter tuning. Though Fu \etal 2014~\cite{Fu_CVPR_2014} performed well in some cases, the approach is not stable as the parameters are preset. Also, benefit from our efficient learning and inference algorithm, we are able to use fully connected hidden nodes to model the rich relations between all objects and humans in the dataset, so that we have more information to detect the common foreground objects.

\begin{wrapfigure}{R}{0.4\textwidth}
\vspace{-0.1in}

  \begin{center}
  \subfigure[]{
  \includegraphics[height = 3.6cm]{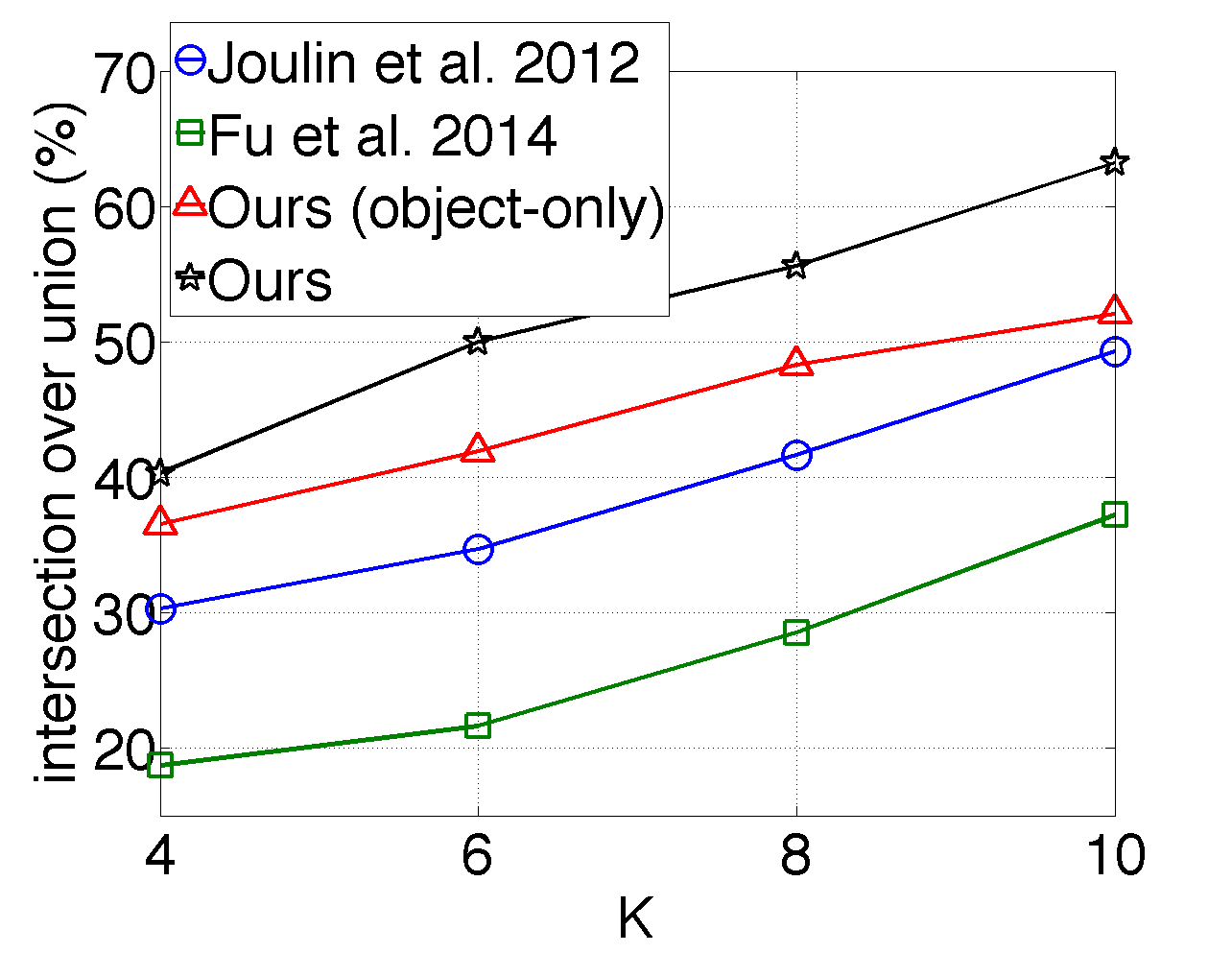}
  \label{fig:class}
  }
\vspace{-0.1in}
  \subfigure[]{
  \includegraphics[height = 3.6cm]{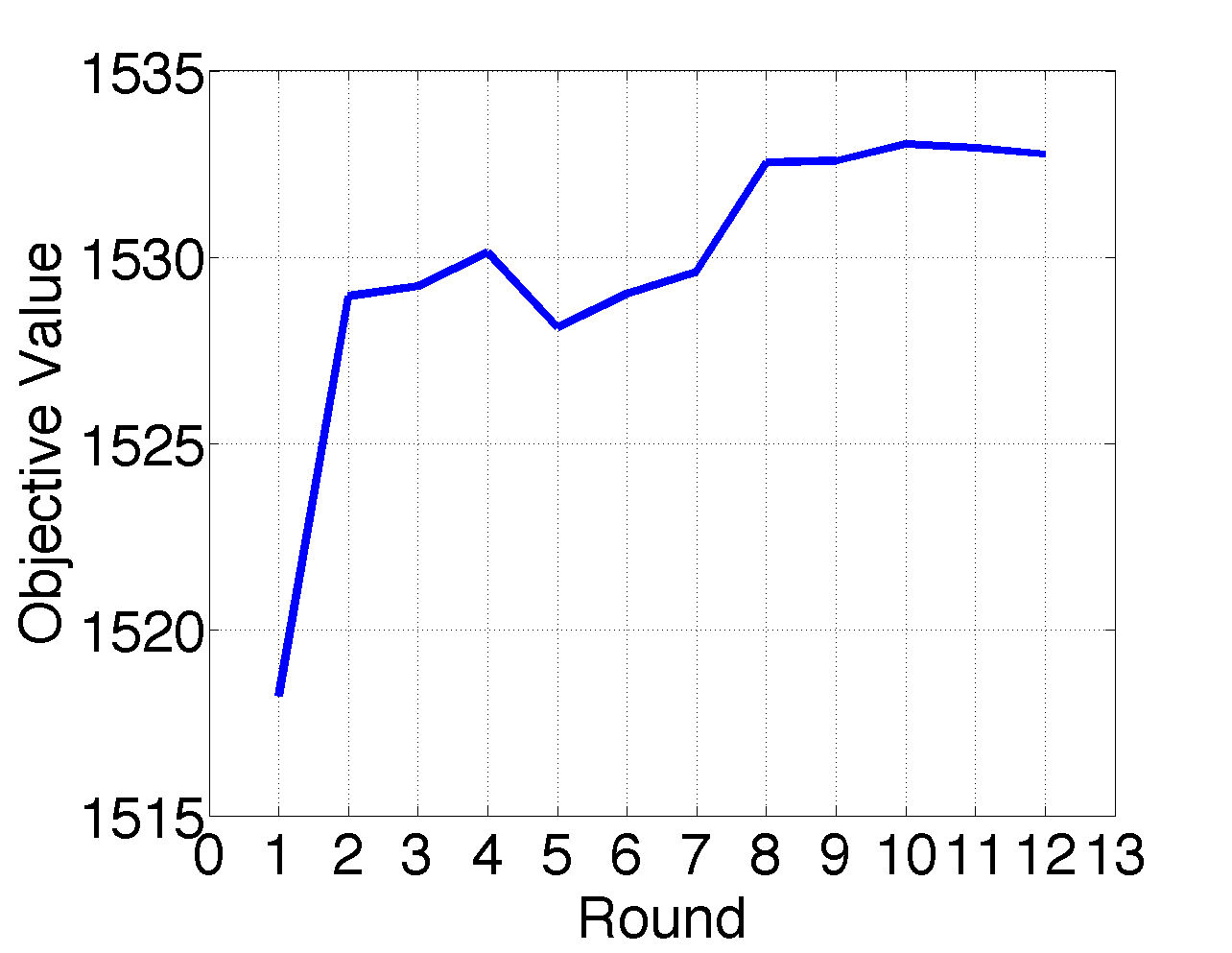}
  \label{fig:cv}
  }
\hspace{-0.2in}
  \caption{(a). Results of table class on Watch-n-Patch dataset varying with cluster number $K$. (b). Learning curve of our approach.}
 \end{center}
\vspace{-0.3in}
\end{wrapfigure}

\header{Can human information be generalized to the segments without humans?}
In our first three datasets with humans in the image, there are a few images where humans are not interacting with the foreground objects such as the fridge on the right in Fig.~\ref{fig:ve_fo}. In these cases, our approach was still possible to segment it out correctly, since it was linked to other visually similar objects which are interacting with humans in the fully connected CRF graph. 

From the results in Table~\ref{tb:re4} on the MS COCO + Watch-n-Patch dataset, we can see that the task is very challenging since images are from different domains with more clustered backgrounds. Though all compared methods perform not more than $20$ percent in accuracy, modeling humans even in a few images still improves the performance.

\header{2D bounding box of human vs. 3D human skeleton.}
From Table~\ref{tb:re3}, we can see that even in 2D images with the bounding box of the detected human, modeling the human-object interactions improves the co-segmentation performance. Using our proposed 3D human-object representation on the more accurate tracked humans gives more improvements as shown in Table~\ref{tb:re1},~\ref{tb:re2}. In the examples in Fig.~\ref{fig:ve_fo}, we also found that our 3D human-object representation is useful to deal with the occlusions and view changes, which is challenging for object visual appearance only based co-segmentation approaches.

\header{How performance varies with cluster number K?}
We show the performance varying with the cluster number $K$ in Fig.~\ref{fig:class}. We can see that the accuracy increase with the class number $K$ as it has higher chance to hit the ground truth regions and more backgrounds are modeled. Our approach has the best performance for each $K$. 

\header{How fast is the learning?}
We also plot a learning curves of our model in Fig.~\ref{fig:cv}. The learning of our approach can be converged mostly within $10$ iterations.

\section{Conclusion and Future Work}\label{sec:con}
In this work, we proposed a novel human centered object co-segmentation approach using a fully connected CRF auto-encoder. We encoded a novel human-object interaction representation and rich object visual features as well as their similarities using the powerful fully connected CRF. Then we used the auto-encoder to learn the parameters from the data itself using an efficient learning even for this complex structure. As a result, we were able to extract those objects which have the most human interactions and are most similar to other objects in the dataset as the foregrounds. In the experiments, we showed that our approach extracted foreground objects more accurately than the state-of-the-art algorithms on two human activity Kinect key frame dataset as well as the musical instruments RGB image dataset. We also showed that our model was able to use the human information in a small portion of images to improve the co-segmentation results. 

In the future, we consider extending our human centered object co-segmentation approach into the semi-supervised setting and incorporating temporal information for video data.


\bibliographystyle{splncs03}
\bibliography{hcoseg}
\end{document}